\documentclass{article}

\usepackage{microtype}
\usepackage{graphicx}
\usepackage{subfigure}
\usepackage{booktabs} 

\usepackage{hyperref}
\usepackage{subcaption}
\usepackage{url}
\usepackage{colortbl}
\usepackage{amscd}
\usepackage{graphicx}
\usepackage{caption}
\usepackage{booktabs}
\usepackage{xcolor}
\usepackage{multirow}

\usepackage[accepted]{icml2024}

\usepackage{amsmath}
\usepackage{amssymb}
\usepackage{mathtools}
\usepackage{amsthm}

\DeclareMathOperator*{\argmax}{arg\,max}

\usepackage[capitalize,noabbrev]{cleveref}

\theoremstyle{plain}

\theoremstyle{definition}

\theoremstyle{remark}

\usepackage[textsize=tiny]{todonotes}

\icmltitlerunning{Image-Caption Encoding for Improving Zero-Shot Generalization}

\begin{document}

\twocolumn[
\icmltitle{Image-Caption Encoding for Improving Zero-Shot Generalization}



\icmlsetsymbol{equal}{*}
\icmlsetsymbol{mitintern}{$\ddagger$}

\begin{icmlauthorlist}
\icmlauthor{Eric Yang Yu}{equal,ucsd,mitintern}
\icmlauthor{Christopher Liao}{equal,bu}
\icmlauthor{Sathvik Ravi}{umd,mitintern}
\icmlauthor{Theodoros Tsiligkaridis}{mitll}
\icmlauthor{Brian Kulis}{bu}
\end{icmlauthorlist}

\icmlaffiliation{mitll}{MIT Lincoln Laboratory}
\icmlaffiliation{ucsd}{University of California San Diego}
\icmlaffiliation{umd}{University of Maryland}
\icmlaffiliation{bu}{Boston University}

\icmlcorrespondingauthor{Eric Yu}{eyyu@ucsd.edu}
\icmlcorrespondingauthor{Christopher Liao}{cliao25@bu.edu}
\icmlcorrespondingauthor{Sathvik Ravi}{sathrav5@umd.edu}
\icmlcorrespondingauthor{Theodoros Tsiligkaridis}{ttsili@ll.mit.edu}
\icmlcorrespondingauthor{Brian Kulis}{bkulis@bu.edu}

\icmlkeywords{Machine Learning, ICML}

\vskip 0.3in
]



\printAffiliationsAndNotice{\icmlEqualContribution \textsuperscript{$\ddagger$} Based on work done at MIT Lincoln Lab} 

\begin{abstract}
Recent advances in vision-language models have combined contrastive approaches with generative methods to achieve state-of-the-art (SOTA) on downstream inference tasks like zero-shot image classification. 
However, a persistent issue of these models for image classification is their out-of-distribution (OOD) generalization capabilities.
We first show that when an OOD datapoint is misclassified, the correct class can be typically found in the Top-$K$ predicted classes.
In order to steer the model prediction toward the correct class within the top predicted classes, we propose the \emph{Image-Caption Encoding (ICE)} method, a straightforward approach that directly enforces consistency between the image-conditioned and caption-conditioned predictions at evaluation time only.
Intuitively, we take advantage of unique properties of the generated captions to guide our local search for the correct class label within the Top-$K$ predicted classes.
We show that our method can be easily combined with other SOTA methods to enhance Top-1 OOD accuracies by \emph{0.5\% on average} and \emph{up to 3\% on challenging datasets}.
Our code: \url{https://github.com/Chris210634/ice}
\end{abstract}


\section{Introduction}

\begin{figure}[h!]
    \centering
    \includegraphics[width=0.5\textwidth]{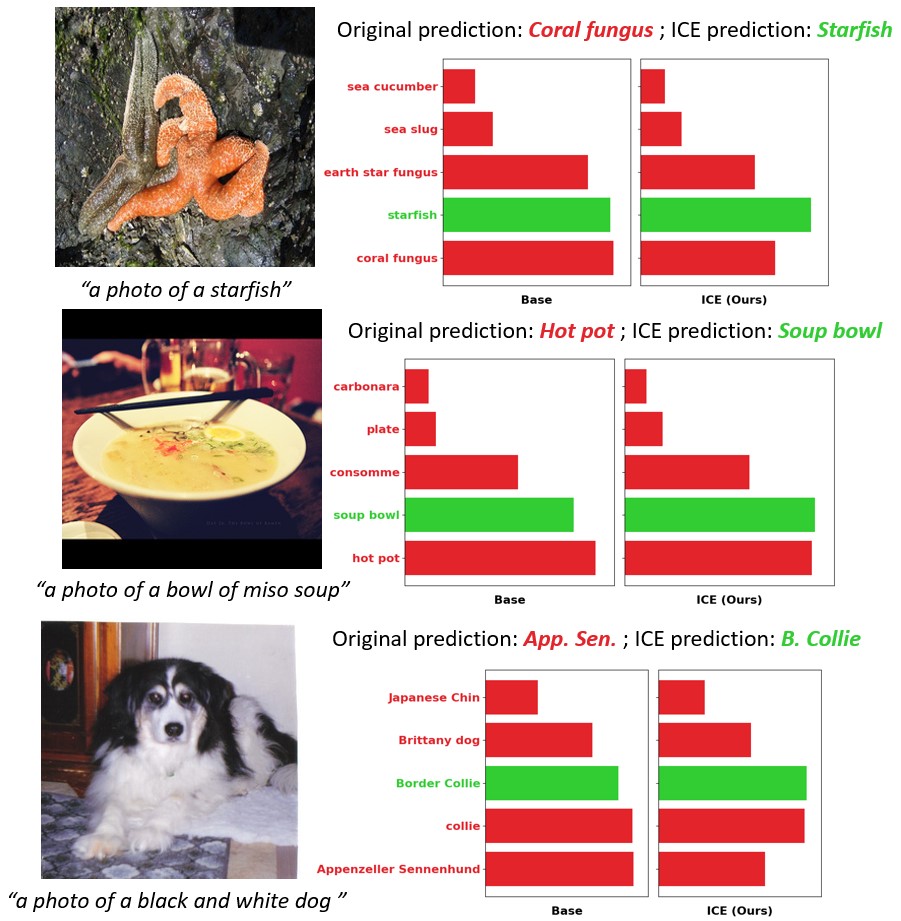}
    \caption{A demonstration for how our ICE method can be used to reclassify correctly. In these examples, ICE is applied directly to a frozen pre-trained CLIP-based model for zero-shot classification. Using the contexts given from the generated captions, ICE is able to successfully influence the pretrained model into predicting the correct classes.}
    \label{fig::ice_01}
\end{figure}

There has been rapid progress in zero-shot image classification over the past two years, thanks to advancements in vision-language (VL) pre-training such as CLIP, ALIGN, and BLIP \citep{radford2021learning, cohen1997align, li2022blip}.
At a high level, these models use a pair of encoders that project visual and textual inputs into a joint latent embedding space. 
As described in the CLIP framework \citep{radford2021learning}, zero-shot classification can be reformulated as an image-to-text retrieval problem, where the class name closest to the image in embedding space is predicted as the label. 
However, state-of-the-art (SOTA) zero-shot classification lags behind in-distribution supervised fine-tuning on all benchmarks. 
In many applications, in-distribution data is not available during training, so fine-tuning on out-of-distribution (OOD) source data in a way that generalizes to unseen data and labels remains a challenging problem. 
Prior works in this area follow two general directions: (1) \emph{Few-shot OOD} methods such as CoOp \citep{zhou2022coop}, CoCoOp \citep{zhou2022conditional}, and MaPLe \citep{khattak2023maple} fine-tune the VL model on generic few-shot source data (e.g. ImageNet). The fine-tuning process is constrained to a carefully selected subset of parameters to ensure generalization to target datasets. (2) \emph{Zero-shot} methods, such as \citet{menon2022visual} and manual prompt ensembling \citep{radford2021learning}, focus on refining the zero-shot prediction without additional fine-tuning. 
These methods do not require additional data, but they typically either require a large closed-source LLM or human-engineered prompts. 

\begin{figure}
    \centering
    \includegraphics[width=0.4\textwidth]{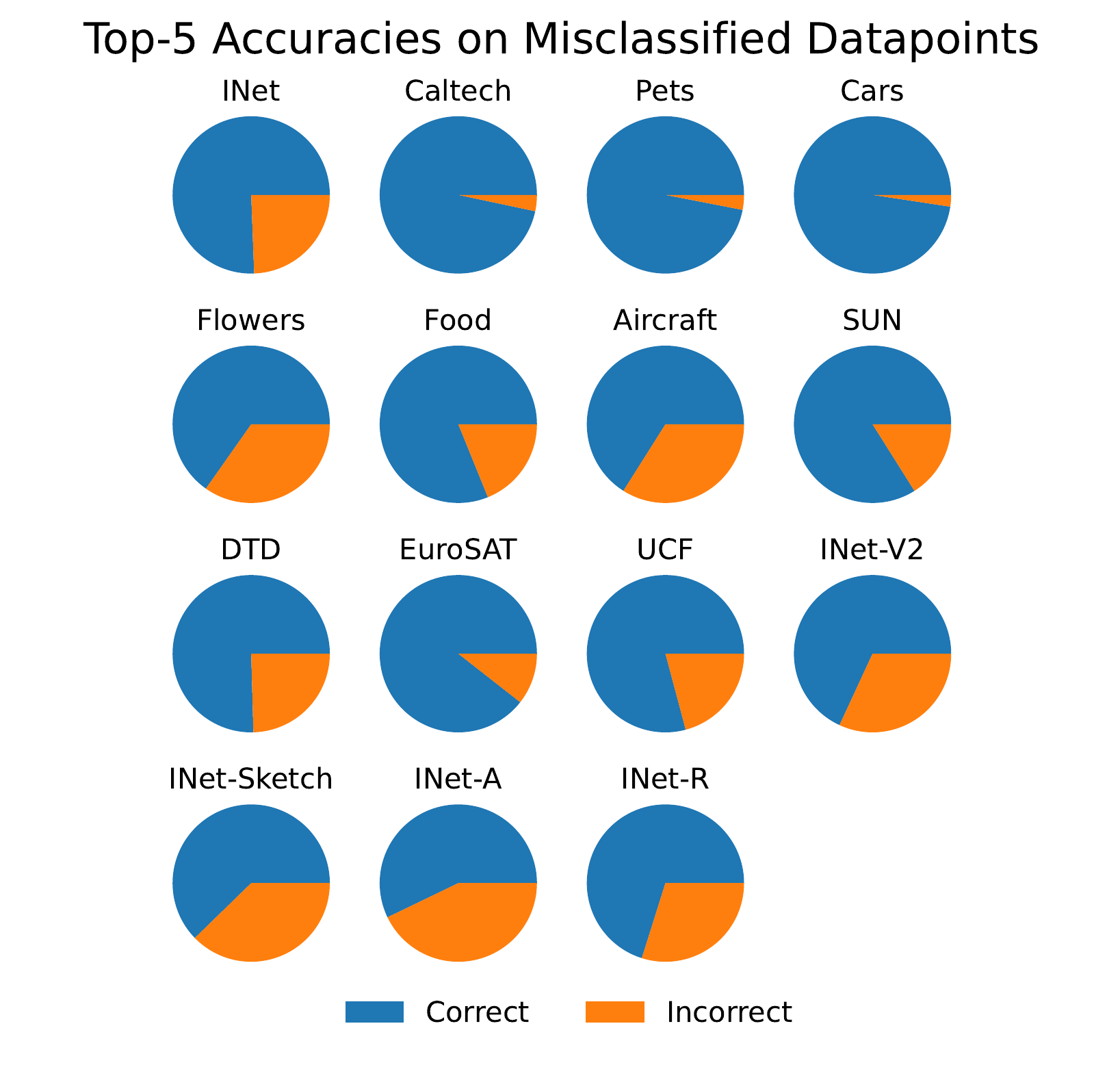}
    \caption{A visualization of the Top-5 accuracies on misclassified Top-1 datapoints in each test dataset. Recall that correct Top-5 classifications form a strict superset over the correct Top-1 classifications. We observe that across all datasets, the true correct class can be found within the Top-5 predicted classes for most misclassified datapoints.}
    \label{fig::mislabels_correct_in_top_5}
\end{figure}

\emph{Our goal is improve zero-shot classification by leveraging captioners, which is previously under-explored.} Towards this goal, we first observe in Figure \ref{fig::mislabels_correct_in_top_5} that the Top-$K$ accuracy (the percentage of samples where the correct label is within the $K$ classes with highest predicted scores, $K > 1$) is consistently higher than the Top-$1$ accuracy.
The reason is that the Top-$K$ predicted classes form a strict superset of the Top-$1$ predicted classes when $K > 1$, and thus characterize a wider range of potentially correct classes.
We observe from Figure \ref{fig::mislabels_correct_in_top_5} that when an image is misclassified, the correct class can usually be found within the Top-$5$ predicted classes. 
Thus, our motivating question is: in order to improve Top-$1$ accuracy, \emph{ how can we steer the model prediction toward the correct prediction within the Top-$K$ predicted classes using generated captions?}

\begin{figure}
    \centering
    \includegraphics[width=0.4\textwidth]{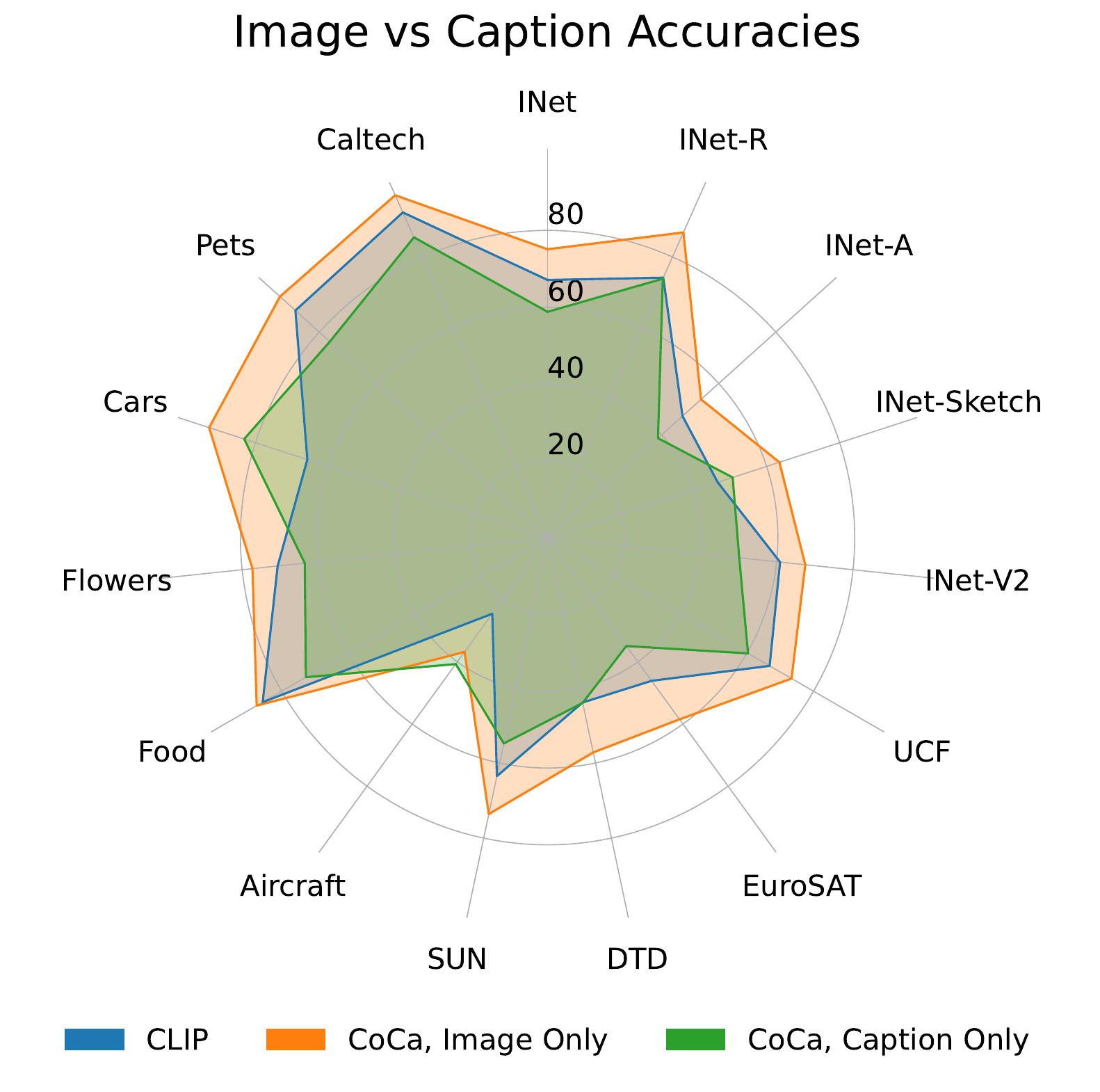}
    \caption{A visualization of Top-1 accuracies between CLIP, CoCa using image embeddings only, and CoCa using caption embeddings only. We observe that while caption embeddings generally underperform compared to standard CoCa, they still retain competitive performance. We include more details on datasets and experiments in Section \ref{subsec::experiments::datasets}.}
    \label{fig::image_vs_caption}
\end{figure}

Before we address this question, we first note that current SOTA zero-shot image classification methods perform a nearest-neighbor search between image and text CLIP embeddings \citep{radford2021learning}.
Recently, many works such as CoCa \citep{yu2022coca}, BLIP-2 \citep{li2023blip} and LLaVA \citep{liu2023visual} extend CLIP with an additional text decoder.
This text decoder is trained to output a description of the image by cross-attending to \emph{all} image tokens outputted by the image encoder. Consequently, the decoder output captures fine-grained spatial information that may be absent from the image cls token. 
Furthermore, the caption verbalizes the content of the image as discrete text tokens, which can oftentimes be used to directly infer the image label.
These advantages are illustrated in Figure \ref{fig::image_vs_caption}, where we use a spider plot to compare the Top-$1$ zero-shot accuracies achieved by CLIP image embeddings, CoCa image embeddings, and CoCa caption embeddings, across 15 datasets.
We notice that while the caption-only CoCa under-performs compared to standard CoCa, it is still competitive and even surpasses CLIP on many datasets.
This suggests that the captions likely contain enough information about the image to supplement the standard zero-shot prediction.

\begin{figure*}[t!]
    \centering
    \includegraphics[width=1\textwidth]{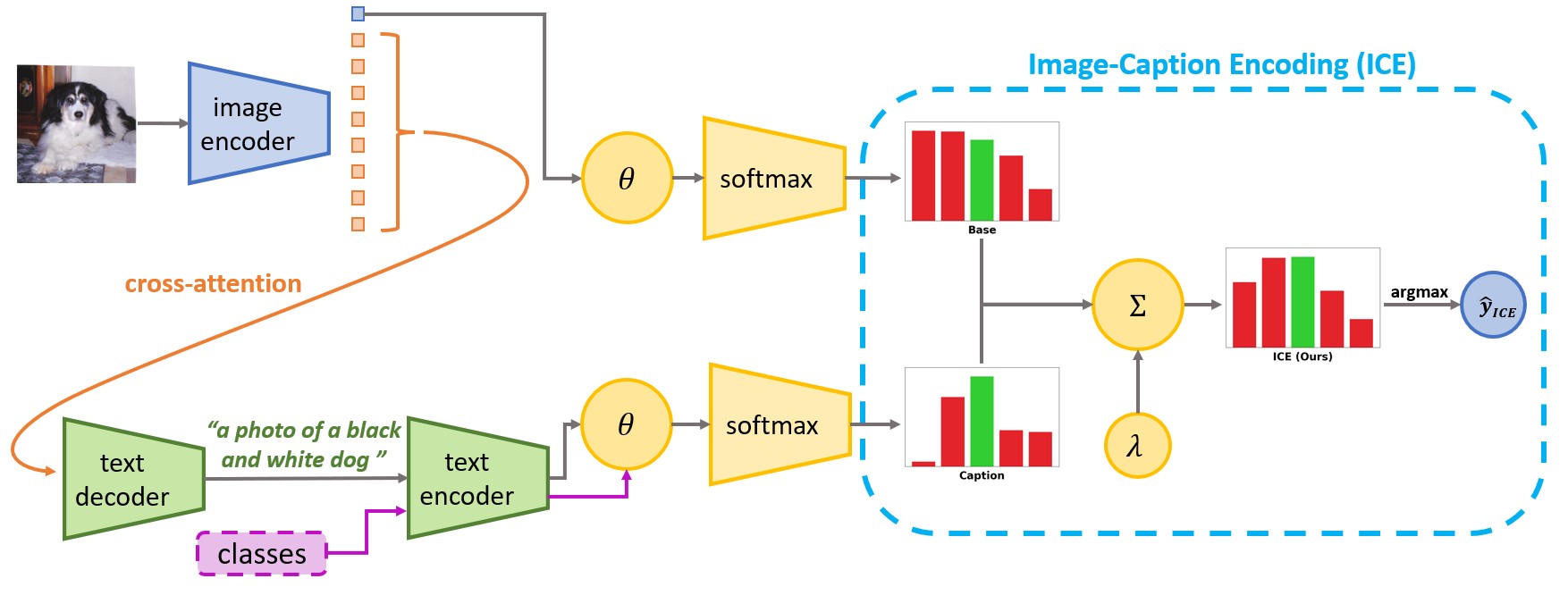}
    \caption{An overview of our \emph{Image-Caption Encoding (ICE)} method. Here, we query a captioner and obtain the caption embedding using the text encoder. We calculate the image and caption probability distributions over the classes by passing the image embeddings, caption embeddings, and class embeddings through the cosine similarity function $\theta$ and softmax operation. Then, we select the Top-$K$ classes and perform a weighted sum of the image and caption probabilities. The weight on the caption prediction $\lambda$ is adaptively selected based on the relative confidence of the image and caption predictions.}
    \label{fig::ice_overview_1}
\end{figure*}

To leverage this additional information,  we propose a novel zero-shot method called \emph{Image-Caption Encoding (ICE)}, where we  combine the information encoded by both image embeddings and caption embeddings in order to make a more informed decision at test time. As illustrated in Figure \ref{fig::ice_overview_1}, ICE is a zero-shot method with no training component and can be easily paired with existing SOTA methods for improved downstream classification performance. 

Although ICE draws inspiration from traditional ensembling techniques, there are several key differences. 
First, ICE leverages the predictions obtained from a \emph{single} model rather than those from several \emph{different} models. 
Second, instead of aggregating predictions over all classes, we only consider the Top-$K$ image predicted classes. 
Third, we incorporate a lightweight confidence selection mechanism that sets the weight on the caption prediction dynamically. 
Finally and most importantly, we exploit specific properties induced within captions that are not present in the image embeddings for standard zero-shot classification. 
We discuss these properties in-depth in Section \ref{subsec::methodology::caption_properties} and demonstrate specific examples of their non-trivial impact in Section \ref{subsec::experiments::analysis_on_ice}.


Our contributions are as follows: 

\noindent $1.$ We extend the zero-shot classification literature by leveraging captioners, which is a direction that is previously under-explored. 

\noindent $2.$ We propose \emph{Image-Caption Encoding (ICE)}, a novel zero-shot classification method that utilizes information from both images and captions to make a decision at evaluation-time only.

\noindent $3.$ We provide experimental results for ICE paired with several different SOTA baselines across 15 different OOD datasets. We show consistent improvements of $0.5\%$ on average and up to $3\%$ on several challenging datasets.

\noindent $4.$ We analyze the benefits and drawbacks of using ICE, and provide ablation studies to analyze the effects of changing different parameters in our ICE method.

\section{Related Works}
\noindent \textbf{Multimodal foundational models}. Many VL foundational models have emerged over the past two years, including CLIP \citep{radford2021learning}, ALIGN \citep{cohen1997align}, BLIP \citep{li2022blip}, CoCa \citep{yu2022coca}, BLIP-2 \citep{li2023blip}, and LLaVA \citep{liu2023visual}. 
These models achieve SOTA on VL tasks by using vast quantities of un-curated image-text data from the web. CLIP uses an image encoder and a text encoder to project the two modalities into a joint latent embedding space. 
Popular downstream applications include zero or few-shot classification and image-text retrieval. 
The more recent models  such as CoCa, BLIP-2 and LLaVA improve CLIP by additionally training a text decoder to explain the embedding space with a caption. 
In the current work, we leverage this additional captioning capability to improve the CLIP zero-shot accuracy. 

\noindent \textbf{Robust fine-tuning}. There is growing interest in fine-tuning multimodal foundational models on limited training data such that the resulting model achieves high accuracy even on domain-shifted data and data with labels not seen during training. 
Many modern approaches rely on prompt tuning and ensembling. 
CoOp \citep{zhou2022coop} is a seminal work which treats the prompt preceding the label names as soft learnable tokens. 
CoCoOp \citep{zhou2022conditional} trains a meta-network to condition the prompt tokens on the image embedding. 
MaPLe \citep{khattak2023maple} shows that learning a conditional visual prompt jointly with textual prompts improves target accuracy. 
All three works achieve impressive results on a diverse set of test datasets despite only being trained on few-shot ImageNet data. 
ClipOOD \citep{shu2023clipood} uses an adaptive margin loss to optimize the visual encoder only, attaining good results on domain generalization benchmarks. 
Our proposed method ICE is a training-free approach that can be readily combined with the above fine-tuning methods to yield higher accuracy on most target datasets.

\noindent \textbf{Ensembling for robust classification}. Ensembling methods leverage multiple diverse predictions to form a robust final prediction; many recent works, including ours, focus on discovering new sources of diversification. 
WiSE-FT \citep{wortsman2022robust} calculates a weight space ensemble of the fine-tuned and pre-trained models to increase robustness under distribution shifts in target data, while inference time remains the same. \citet{lowell2023ftswa} demonstrate that combining cross-entropy fine-tuning with stochastic weight averaging improves domain generalization. \citet{menon2022visual} use GPT descriptions to generate a more diverse set of text prototypes for zero-shot classification. 
\citet{radford2021learning} use an ensemble of 80 handcrafted prompts to achieve the same goal.
In our paper, while drawing inspiration from ensembling, we introduce the novel use of captions as a unique source of diversification, an avenue not yet explored by previous studies. Importantly, our approach is designed to seamlessly integrate with other zero-shot methods, as demonstrated in our experimental results.

\section{Methodology}
\subsection{Preliminaries}
Consider a dataset $\mathcal{D} \subset \mathcal{I} \times \mathcal{T}$ where $\mathcal{I}$ is the image domain and $\mathcal{T}$ is the text domain, and $(I_i, T_i)$ forms a corresponding image-text pair (i.e. $T_i$ is a caption that describes $I_i$). 
In the CLIP framework~\citep{radford2021learning}, there are two main architectural components: an image encoder $f_\mathcal{I}: \mathcal{I} \rightarrow \mathbb{R}^l$ that maps images to a shared latent space, and a text encoder $f_\mathcal{T}: \mathcal{T} \rightarrow \mathbb{R}^l$ that maps text to the same shared latent space. 
Both encoders are pre-trained using a contrastive loss that pulls corresponding image-text embeddings close together in latent space, and pushes non-corresponding image-text embeddings away from each other. 
In our framework, we require an additional text decoder, pre-trained using a next-token-prediction loss, to provide captioning functionality. 

In image classification following the CLIP framework~\citep{radford2021learning}, for an image vector $I$ and class labels vector $y := [y_1, y_2, \dots, y_m]^\intercal$, we first feed each class label through a prompt skeleton to obtain class prompts (e.g. for class ``cat'' and a prompt skeleton ``A photo of a \{\}'', the resulting prompt is ``A photo of a cat''). Then, both the image and class prompt vectors pass through the image and text encoders to obtain latent embeddings $\widetilde{I}$ and $\widetilde{y}$, respectively. The predicted class label $\hat{y}$ is then given by $\argmax_i \theta (\widetilde{I}, \widetilde{y}_i), i \in \{1, 2, \dots, m\}$, where $\theta: \mathbb{R}^{l} \times \mathbb{R}^{l} \rightarrow \mathbb{R}$ is the cosine similarity function.

\begin{figure*}[t!]
    \centering
    \includegraphics[width=1\textwidth]{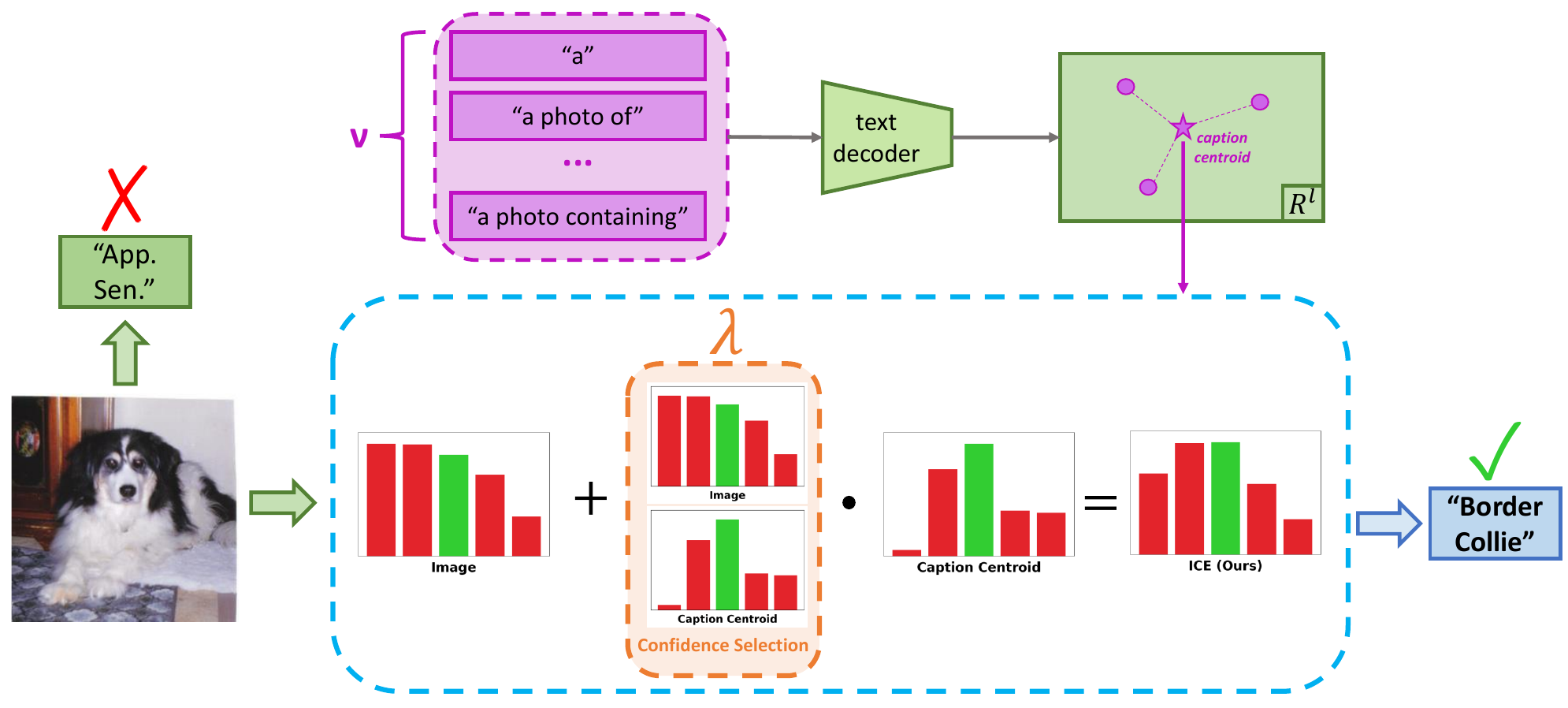}
    \caption{A more detailed look at how our \emph{Image-Caption Encoding (ICE)} method works. In practice, instead of using a single caption for ICE, we use the centroid of $\upsilon$ differently-prompted caption embeddings. Then, using the centroid caption embedding, we adaptively select the $\lambda$ weight by comparing the standard deviations of the image prediction probabilities and caption prediction probabilities, over the Top-$5$ classes. The final ICE scores are then a $\lambda$-weighted sum between the two probability distributions.}
    \label{fig::ice_overview_2}
\end{figure*}

\subsection{Image-Caption Encoding}

Consider a text decoder $f_\mathcal{\phi}: \mathcal{P} \times \mathcal{I} \rightarrow \mathcal{T}$ that maps a prompt $p \in \mathcal{P}$ (e.g. "a photo of") and an image $I$ to a caption $c \in \mathcal{T}$. We can feed caption $c$ back through the text encoder to obtain $\widetilde{c} := f_\mathcal{T}(c)$.

Using the Softmax function \citep{bridle1989softmax}, we obtain the class probabilities for image $I$ and caption $c$, respectively, as 

\begin{equation}
    \label{eq::image_caption_similarity_scores}
    \begin{aligned}
        &S^I := \text{Softmax}\biggl(\Bigl[ \theta(\widetilde{I}, \widetilde{y}_1), \theta(\widetilde{I}, \widetilde{y}_2), \dots, \theta(\widetilde{I}, \widetilde{y}_m) \Bigr]^\intercal\biggr) \\
        &S^c := \text{Softmax}\biggl(\Bigl[ \theta(\widetilde{c}, \widetilde{y}_1), \theta(\widetilde{c}, \widetilde{y}_2), \dots, \theta(\widetilde{c}, \widetilde{y}_m) \Bigr]^\intercal\biggr).
    \end{aligned}
\end{equation}

The indices corresponding to the $K$ classes with highest image-predicted probability, and the final ICE prediction, are computed respectively as:

\begin{equation}
    \label{eq::top_k_and_ice_prediction}
    \begin{aligned}
        &\Omega^I_K := \argmax_{J \subset M, |J| = K } \sum_{j \in J} S^I_j 
        &\argmax_{\omega \in \Omega^I_K} S^I_\omega + \lambda S^c_\omega
    \end{aligned} 
\end{equation} \\
where $M = \{1, 2, \dots, m\}$ and $\lambda$ is a scalar variable. 
In essence, the ICE prediction remains anchored within the primary Top-$K$ classifications as determined by the image class probabilities. When we incorporate the caption scores tied to these Top-$K$ predictions, it reshapes the probability landscape over the initial Top-$K$ image-determined classes.
By selecting the class with the highest probability from this refined distribution, we aim to align closer with the true class while retaining the accuracy of previous classifications. 
Intuitively, the caption probability distribution should provide information about the image that is not clear or fully captured by the image probability distribution, as detailed in Section \ref{subsec::methodology::caption_properties}, and their aggregated prediction should provide a more reasonable downstream prediction. Figures \ref{fig::ice_overview_1} and \ref{fig::ice_overview_2} contain a high-level overview and an in-depth visual interpretation of our method, respectively.

\subsection{Additional Modifications}
As visualized in Figure \ref{fig::ice_overview_2}, in our experiments, we use the centroid of a diverse set of captions rather than a single caption for increased robustness. 
That is, for a set of prompts $P \in \mathcal{P}^\upsilon$, we generate a set of corresponding captions $C \in \mathcal{T}^\upsilon$, obtain their caption embeddings $\widetilde{C} := \{f_\mathcal{T}(c_1), f_\mathcal{T}(c_2), \dots f_\mathcal{T}(c_\upsilon)\}$, and finally, their centroid $\widetilde{\overline{c}} := \frac{1}{\upsilon}\sum^\upsilon_i \widetilde{c}_i$. The centroid $\widetilde{\overline{c}}$ is then used in place of $c$ in Equations \ref{eq::image_caption_similarity_scores} and \ref{eq::top_k_and_ice_prediction}.

In addition, we dynamically compute the caption scalar variable $\lambda \in \mathbb{R}_+$ as a function of the standard deviation of the captions. That is, given some image $I$ and caption $c$, we compute $\lambda$ as
\begin{equation}
    \label{eq::lambda}
    \begin{aligned}
        \lambda = \xi \frac{\sigma(S^c_K)}{\max(||[\sigma(S^I_K), \sigma(S^c_K)]||_2, \epsilon)}
    \end{aligned}
\end{equation}
where $\xi$ is a constant, $\epsilon$ is a small constant, $S^I_K$ and $S^c_K$ are the Top-$K$ probabilities for $S^I$ and $S^c$ respectively (i.e. the probabilities whose indices are specified by $\Omega^I_K$), and $\sigma$ is the standard deviation operator. 
Intuitively, the standard deviation of the Top-$K$ highest probabilities of the image and caption distributions correspond to the model confidence about their respective predictions.
On one hand, when the model confidence for the image prediction is high and the caption confidence is low, then the caption probabilities should not influence the image probabilities as much.
On the other hand, when the image confidence is low but the caption confidence is high, the caption probabilities should more heavily influence the image probabilities.
In the event when both image and caption confidences are high or low, the default weighting would be relatively equal.
The constant $\xi$ can specify how much the caption probabilities should affect the image probabilities overall.

\subsection{Caption Properties} 
\label{subsec::methodology::caption_properties}

In general, the zero-shot accuracy obtained using only the caption embeddings is significantly (on average about 5 \%, see Table \ref{tab:zs}) lower than the zero-shot accuracy obtained using image embeddings, with the notable exception of aircraft fine-grained classification. 
Caption-only zero-shot classification is often unreliable, since the caption does not always correspond to one of the label choices. 
For example, a picture containing a teddy bear on top of a bed might be captioned as ``a picture of a teddy bear'', ignoring the bed in the background. 
However, if teddy-bear is not one of the labels, and the correct label is ``bed'', the caption does not provide useful information for the classification problem. 
For this reason, the optimal hyperparameters for ICE place a greater emphasis on the prediction from the image embedding $\widetilde{I}$.
Nonetheless, for the caption embedding to contribute to a higher aggregate accuracy, we only require that the caption-predicted probabilities $S^c_K$ be not completely correlated with $S^I_K$. 
In other words, \emph{the caption sometimes contains extra information that nudges the prediction in the correct direction}. 
We list here a few intuitions for why using captions can improve overall classification accuracy:

\noindent $1.$ The text decoder cross-attends to all output image tokens from the vision encoder, while the image prediction only uses the output cls token. The image token matrix contains spatial information that may be pertinent to the target task.

\noindent $2.$ The text decoder was pretrained to caption images with a language-modeling loss. Consequently, it exhibits some rudimentary reasoning ability based on learning relationships between certain concepts. For example, the text decoder has learned that the painting ``the starry night'' is authored by Vincent van Gogh. This correspondence is learnt by the weights of the decoder and may be useful for some classification tasks. In our experiments, we found that the caption prediction is much better than the image prediction on aircraft classification. This is likely because the correspondence between fine-grained visual concepts and the aircraft model name is learned by the text decoder.

\noindent $3.$ The caption effectively isolates visual concepts; this is an inherent property of textual data. For example, a caption that reads ``a rough red blanket'' effectively isolates the texture, color and content of the image. In our experiments, we found that captions on the EuroSAT dataset often isolate the land-use information from the geographical information, e.g. ``a photo of agricultural land in China''. The caption explicitly separates useful information (agricultural land) from information that is irrelevant to the classification problem (China). Consequently, a classifier trained on captions is less likely to learn domain-specific spurious correlations, especially in the few-shot setting.

We provide concrete examples of these discussed intuitions in our empirical analysis in Section \ref{subsec::experiments::analysis_on_ice}.

\section{Experiments}


In our experiments in Table \ref{tab:zs}, we analyze the impact of combining ICE with different zero-shot (ZS) baselines across a suite of benchmarks. 
We show that our method consistently improves ZS baselines without requiring additional training. 
In addition, we analyze several data points to show how ICE improves over the base method that it is paired with.
All implementation details can be found in Appendix \ref{sec::exp_imp_details}.
To demonstrate the general applicability of our ICE method, we also present results using three different recent multimodal models that contain a decoder component: CoCa \citep{yu2022coca}, BLIP-2 \citep{li2023blip}, and LLaVA \citep{liu2023visual}. These results are included in Table \ref{tab:zs-appendix} in the Appendix.

\paragraph{Datasets.} \label{subsec::experiments::datasets}
In line with prior work, our datasets are split into two categories: \emph{cross-dataset evaluation} and \emph{domain generalization}. For cross-dataset generalization, each evaluation dataset has mostly non-overlapping classes and unrelated data distributions for zero-shot classification. For domain generalization, the evaluation datasets are domain-shifted variations of the ImageNet dataset and share the same classes as ImageNet. We evaluate our method on 11 cross-dataset generalization datasets covering a wide range of image recognition tasks. These include two generic objects datasets, ImageNet \citep{russakovsky2014imagenet} and Caltech101 \citep{li2004caltech101};
five fine-grained datasets, OxfordPets \citep{parkhi2012oxfordpets}, StanfordCars \citep{krause2013stanfordcars}, Flowers102 \citep{nilsback2008flowers102}), Food101 \citep{bossard2014food101}, and FGVCAircraft \citep{maji2013fgvcaircraft};
a scene categorization dataset, SUN397 \citep{xiao2010sun397};
an action recognition dataset, UCF101 \citep{soomro2012ucf101};
a describable textures dataset, DTD \citep{cimpoi2013dtd},
and a satellite images dataset, EuroSAT \citep{helber2017eurosat}. In addition, we consider four domain generalization datasets, each applying a different distribution shift to the source ImageNet dataset. 
These include an extension of the ImageNet dataset, ImageNetV2 \citep{recht2019imagenetv2};
a black and white sketches dataset, ImageNet-Sketch \citep{wang2019imagenetsketch},
a naturally adversarial dataset, ImageNet-A \citep{hendrycks2019imageneta},
and a dataset containing different renditions (e.g. cartoons, graffiti, plush objects, etc.) of the ImageNet classes, ImageNet-R \citep{hendrycks2020imagenetr}.

\subsection{Zero-Shot Classification}

\setlength\tabcolsep{4.7 pt}
\begin{table*}[t!]
\scriptsize
\centering
\begin{tabular}{lccccccccccccccccc}
\toprule
&  & \multicolumn{11}{c}{\textbf{Cross-dataset Evaluation Targets}} & \multicolumn{5}{c}{\textbf{Domain Generalization Targets}} \\
 \cmidrule(lr){3-13} \cmidrule(lr){14-18}

     & \rotatebox{90}{ INet } & \rotatebox{90}{ Caltech } & \rotatebox{90}{ Pets } & \rotatebox{90}{ Cars } & \rotatebox{90}{ Flowers } & \rotatebox{90}{ Food } & \rotatebox{90}{ Aircraft } & \rotatebox{90}{ SUN } & \rotatebox{90}{ DTD } & \rotatebox{90}{ EuroSAT } & \rotatebox{90}{ UCF } & \rotatebox{90}{ Average } & \rotatebox{90}{ INet-V2 } & \rotatebox{90}{ INet-Sketch } & \rotatebox{90}{ INet-A } & \rotatebox{90}{ INet-R } & \rotatebox{90}{ Average }\\
    \midrule

    Zero-shot(Image)  & 75.1 & \textbf{97.6} & \textbf{93.8} & 92.7 & 77.3 & 87.5 & 36.8 & 73.6 & 57.2 & 58.5 & 73.4 & 74.8 & 67.5 & 63.5 & 53.8 & 87.0 & 68.0 \\
    Zero-shot (Caption) & 58.8 & 85.6 & 76.3 & 83.1 & 63.6 & 72.7 & \textbf{40.7} & 54.8 & 44.0 & 34.9 & 60.3 & 61.6 & 50.2 & 50.7 & 38.7 & 73.8 & 53.3 \\
    \rowcolor{lightgray}
      + ICE  & \textbf{75.6} & 97.1 & \textbf{93.8} & \textbf{93.0} & \textbf{78.0} & \textbf{87.7} & {38.3} & \textbf{74.0} & \textbf{59.3} & \textbf{61.3} & \textbf{74.3} & \textbf{75.7} & \textbf{67.8} & \textbf{64.0} & \textbf{54.4} & \textbf{87.5} & \textbf{68.4} \\
    \midrule
    Manual Prompts  & 75.5 & \textbf{97.1} & 93.7 & 92.7 & 77.5 & 87.5 & 37.5 & 74.0 & 60.7 & 60.2 & 73.8 & 75.5 & 67.8 & \textbf{64.7} & 53.2 & 88.1 & 68.4 \\
    \rowcolor{lightgray}
      + ICE  & \textbf{75.9} & \textbf{97.1} & \textbf{93.8} & \textbf{93.1} & \textbf{78.7} & \textbf{87.6} & \textbf{39.7} & \textbf{74.1} & \textbf{61.6} & \textbf{61.2} & \textbf{73.9} & \textbf{76.1} & \textbf{68.2} & \textbf{64.7} & \textbf{54.4} & \textbf{88.4} & \textbf{68.9} \\
    \midrule
    GPT Centroids  & 74.8 & \textbf{97.8} & \textbf{93.3} & 92.4 & 75.8 & 87.4 & 36.4 & 73.9 & 58.8 & 63.7 & 73.2 & 75.3 & 67.3 & 63.3 & 52.6 & 86.7 & 67.5 \\
    \rowcolor{lightgray}
      + ICE  & \textbf{75.2} & 97.4 & 93.0 & \textbf{92.8} & \textbf{76.3} & \textbf{87.7} & \textbf{39.1} & \textbf{74.2} & \textbf{60.2} & \textbf{64.2} & \textbf{73.7} & \textbf{75.9} & \textbf{67.4} & \textbf{63.5} & \textbf{53.5} & \textbf{87.1} & \textbf{67.9} \\
    \midrule
    GPT Score Mean  & 74.9 & \textbf{97.6} & \textbf{93.7} & 92.4 & 76.2 & 87.3 & 36.2 & 73.9 & 58.9 & 64.9 & 73.6 & 75.5 & 67.6 & 63.5 & 52.8 & 86.8 & 67.7 \\
    \rowcolor{lightgray}
      + ICE  & \textbf{75.4} & 97.4 & 93.5 & \textbf{92.8} & \textbf{77.0} & \textbf{87.6} & \textbf{39.2} & \textbf{74.2} & \textbf{60.2} & \textbf{65.5} & \textbf{73.9} & \textbf{76.1} & \textbf{67.9} & \textbf{63.7} & \textbf{53.4} & \textbf{87.2} & \textbf{68.1} \\
    \bottomrule
  \end{tabular}
  \caption{Comparison with zero-shot baselines on 15 test datasets. We always observe that stacking our ICE method on top of baseline methods provides consistent improvements. All methods are zero-shot and use CoCa ViT-L/14. The caption zero-shot accuracy is reported using one caption embedding prompted by ``a photo of''. ImageNet is abbreviated INet.}
  \vspace{-1.5em}
  \label{tab:zs}
\end{table*}
\setlength\tabcolsep{6 pt}

\paragraph{Baselines.}
We consider four existing SOTA methods as zero-shot classification baselines for ICE: (1) a pre-trained CoCa model \citep{yu2022coca} with class embeddings generated using the prompt ``a photo of a \{\}'', where ``\{\}''is replaced by the corresponding class 
(2) the pre-trained model with class embeddings calculated from the centroids of 80 intermediate embeddings generated using hand-crafted prompts from \citet{radford2021learning} 
(3) the pre-trained model using large language model (LLM) generated descriptors from \citep{menon2022visual}
(4) a variation of the previous baseline, where we take the centroid of each descriptor embedding for standard zero-shot classification.


\paragraph{Results.}
In Table \ref{tab:zs}, we observe that stacking ICE achieves consistent improvements of around 0.5\% on average across cross-dataset evaluation and domain-generalization evaluation benchmarks, with improvements of up to 3\% on datasets like FGVCAircraft and EuroSAT. We emphasize that these ZS accuracy improvements are consistent across all datasets (Table \ref{tab:zs}) and three different pretrained model architectures (CoCa, BLIP-2 and LLaVA in Table \ref{tab:zs-appendix}).

\subsection{Few-Shot Classification}

Although our ICE method is a zero-shot method, it can be trivially extended to the few-shot learning setting by applying it to the few-shot finetuned model. Accordingly, we consider SOTA methods where a pre-trained model is fine-tuned on 16-shot ImageNet training data.
Specifically, the ImageNet training dataset contains 1000 classes with 16 images per class, for a total of 16,000 images in the train dataset. These few-shot results are included in Table \ref{tab:ft} of the Appendix.

\paragraph{Baselines.}
We consider two prompt learning methods, CoOp \citep{zhou2022coop} and MaPLe \citep{khattak2023maple}, and a fine-tuning method CLIPood \citep{shu2023clipood} as our ICE baselines. 
We use each method for few-shot fine-tuning on the CLIP components of the CoCa \citep{yu2022coca} architecture, and perform standard zero-shot classification by following the CLIP framework \citep{radford2021learning}.

\paragraph{Results.}
As seen in Table \ref{tab:ft} of the Appendix, applying ICE to each baseline provides improvements of $0.5\%$ on average across cross-dataset generalization evaluation datasets, and smaller improvements for the domain generalization datasets.
For the methods such as CLIPood where we do not see improvements in domain generalization on average, we find that ICE at least maintains the average performance of the classification backbone.

\subsection{Understanding Why ICE Provides Improvements} \label{subsec::experiments::analysis_on_ice}
\begin{figure}[t!]
    \centering
    \includegraphics[width=0.5\textwidth]{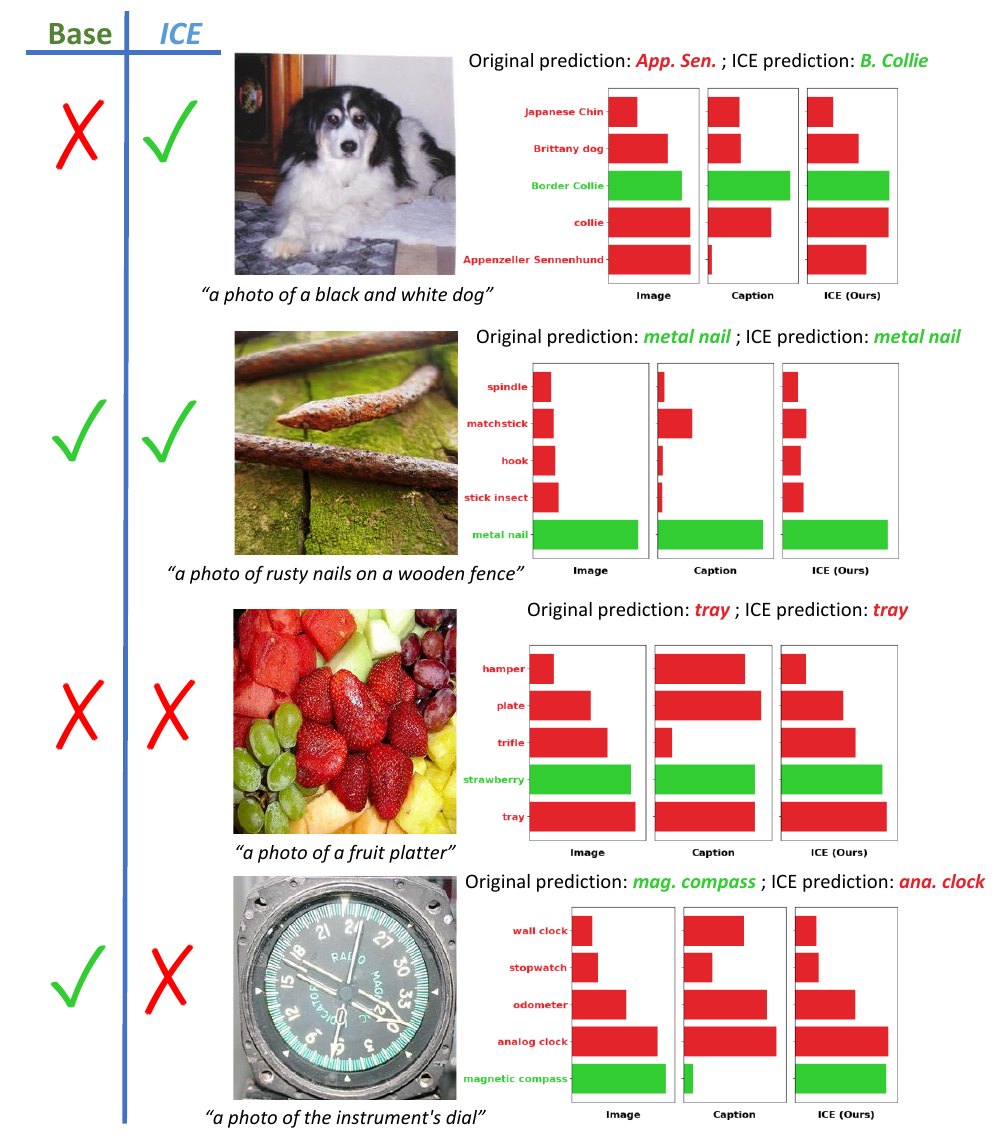}
    \caption{A qualitative analysis on various ways ICE can affect the downstream classification performance.}
    \label{fig::ice_analysis}
\end{figure}

To better understand why ICE improves the base methods paired with it, we analyze examples in Figure \ref{fig::ice_analysis} where ICE correctly reclassifies a previously incorrectly classified datapoint, where ICE preserves a previously correct classification, where ICE fails to correctly reclassify a previously incorrect classification, and where ICE accidentally incorrectly reclassifies a previously correct classification.

\begin{figure*}[t!]
\centering
\begin{subfigure}
  \centering
  \includegraphics[width=0.3\linewidth]{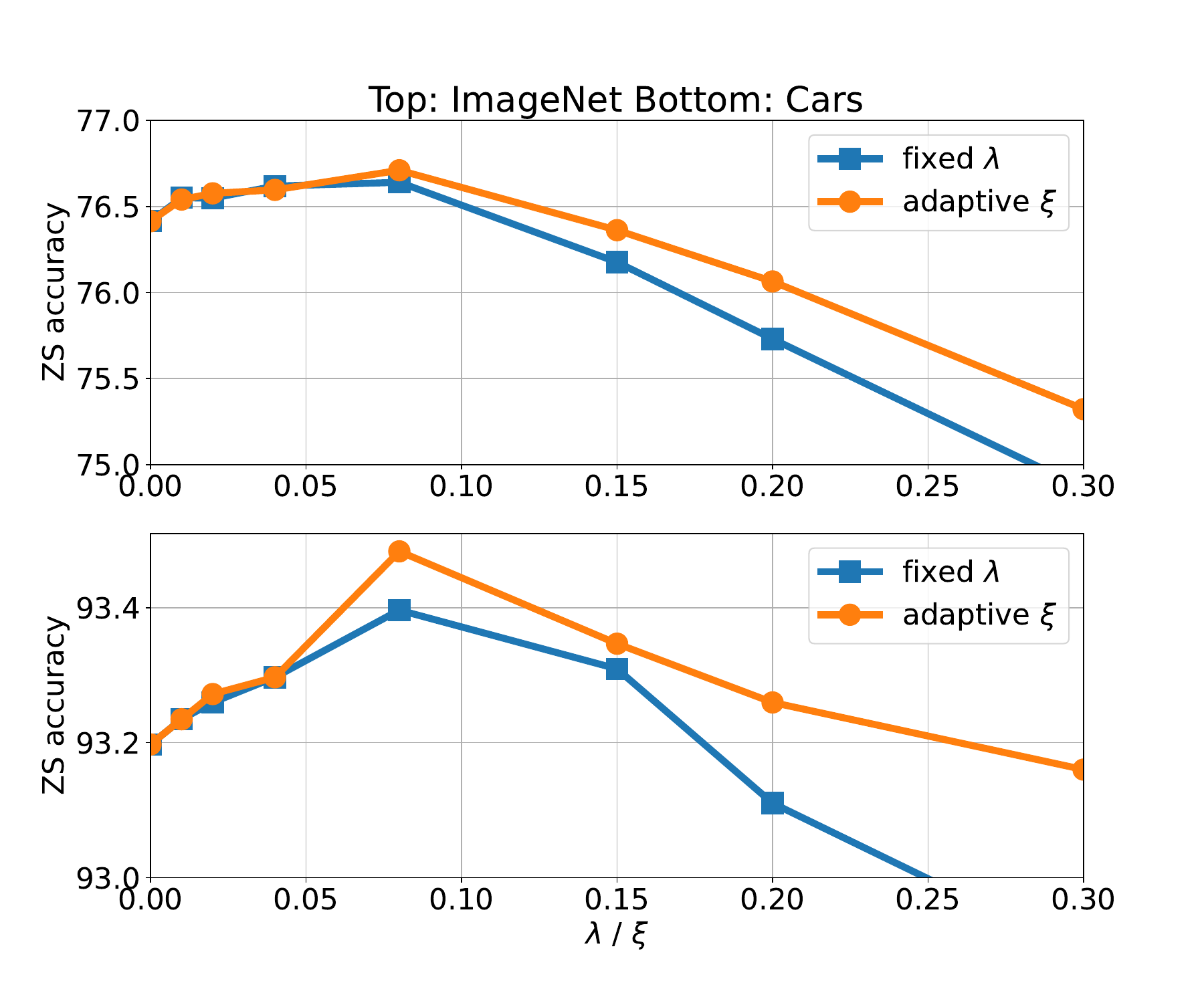}
\end{subfigure}
\begin{subfigure}
  \centering
  \includegraphics[width=0.3\linewidth]{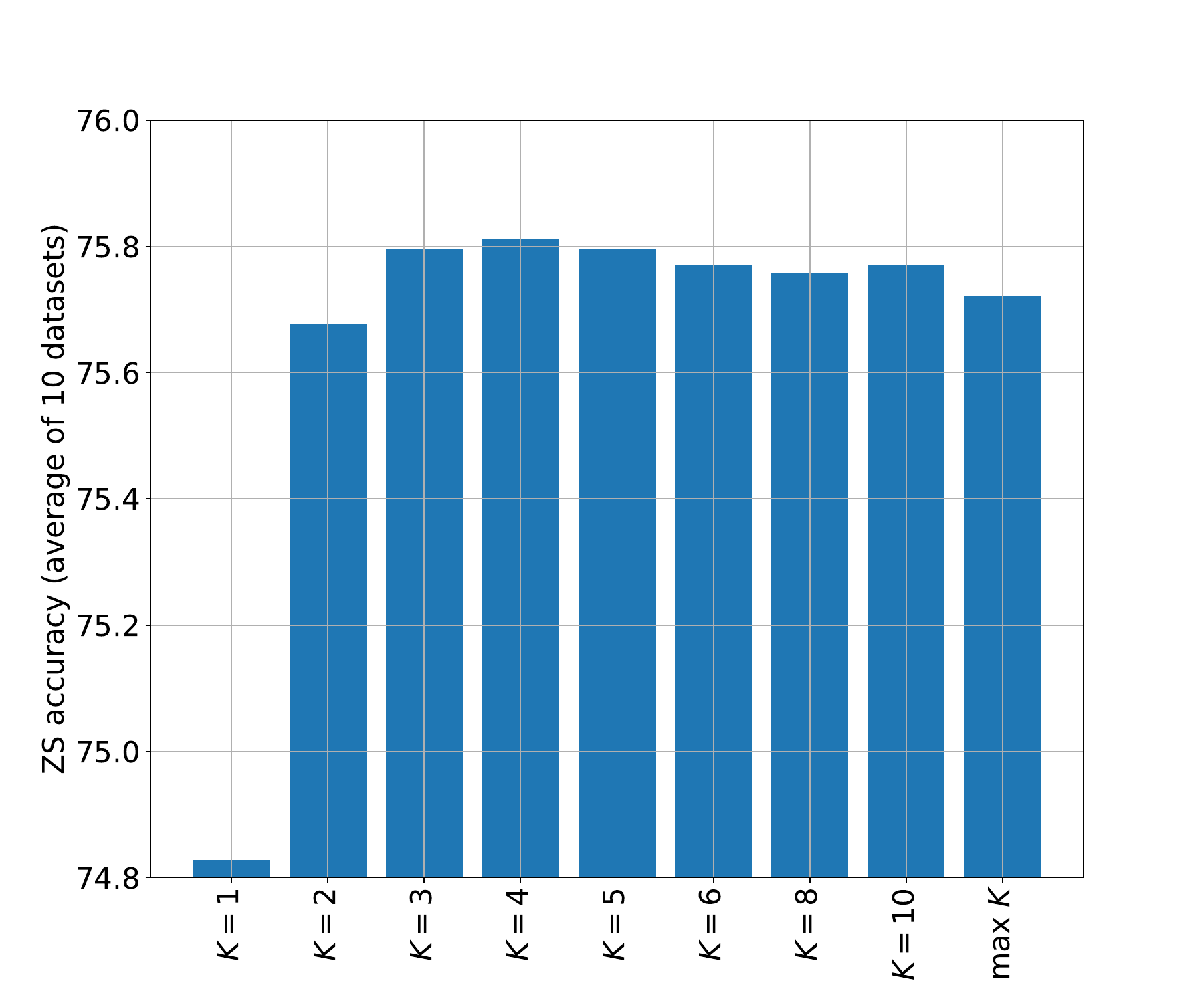}
\end{subfigure}
\begin{subfigure}
  \centering
  \includegraphics[width=0.3\linewidth]{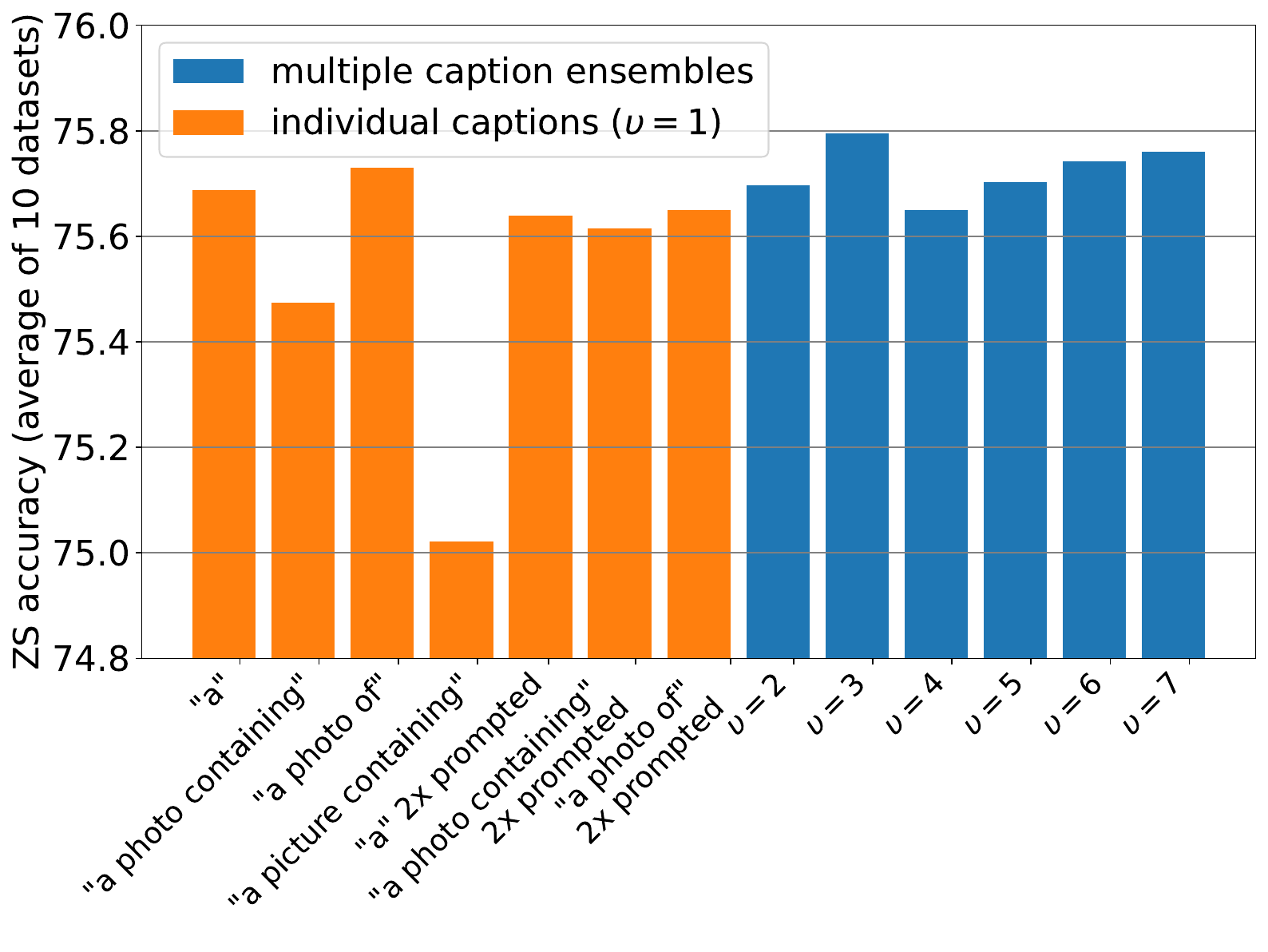}
\end{subfigure}
\caption{Top: Ablation results on varying $\xi$. Comparison between fixed caption weight $\lambda$ and adaptive $\lambda$ using Eq. \ref{eq::lambda}. Adaptive $\lambda$ is clearly superior on Cars and ImageNet. Middle: Ablation results on varying caption prompting and number of captions in ensemble ($\upsilon$). $\upsilon=3$ captions is optimal. Bottom: Ablation results on varying $K$. $K=4$ is optimal and clearly superior than bypassing Top-$K$ selection step (denoted as max $K$ in bar plot). }
\label{fig:ablations}
\end{figure*}

In the first example of Figure \ref{fig::ice_analysis}, we observe that ICE reclassifies the incorrectly predicted \emph{Appenzeller Sennenhund} class to the correct \emph{Border Collie} class. One intuition for why this occurs is that Border Collies most commonly have black and white fur, whereas Appenzeller Sennenhund dogs typically have unique tricolor fur coats. Thus, the caption ``a photo of a black and white dog'' would correspond more with the Border Collie purely based on its bicolor fur. Additional examples of correct ICE reclassifications can be found in Figure \ref{fig::ice_01}.

In the second example, we observe that ICE is able to successfully preserve an originally correct classification. Here, since the caption agrees with the image predicted class, ICE is able to predict the same class as before.

In the third image, ICE fails to correctly reclassify an initially incorrect prediction. In this case, the original prediction was a \emph{tray}, which would make sense since this image can technically be described as a ``tray of fruits''. Similarly, the captions describe the image as ``a photo of a fruit platter'', which can correspond more to a \emph{plate} or \emph{tray} than a \emph{strawberry}. Thus, the ICE predicted class still matches that of the original class.

Finally, the fourth image is an example for how ICE can accidentally incorrectly reclassify an initially correct prediction. Here, the correct class is a \emph{magnetic compass}, as evident by the blue text written on the instrument. While the caption ``a photo of the instrument's dial'' technically describes the image, without additional visual context, it is easy to believe that the caption is describing a clock rather than a magnetic compass since dials tend to be associated with clocks. Due to this ambiguity, ICE was able to convince the initial predicted class to reclassify to the runner-up class of \emph{analog clock}. This example highlights the importance of why we need to consider both the image and caption information for making the most informed decisions.

\section{Ablation Studies}
A comprehensive ablation study on the parameters of ICE is presented in Figure \ref{fig:ablations}. First, we evaluate the contribution of the adaptive $\lambda$ mechanism proposed in Eq. \ref{eq::lambda} on the left of the figure. On ImageNet and cars, the zero-shot accuracy of the adaptive $\lambda$ is clearly superior to the fixed $\lambda$ for varying values of $\xi$ (in the adaptive case) and $\lambda$ (in the fixed case). In the middle bar plot, we examine the contribution of the Top-$K$ selection procedure outlined in Eq. \ref{eq::top_k_and_ice_prediction}. We compare multiple values of $K$ (where $K=1$ is equivalent to ignoring caption embeddings, and ``max $K$'' denotes score averaging over all classes without Top-$K$ selection). Values of $K$ around $K=4$ are clearly superior to both ignoring caption embeddings and score averaging without Top-$K$ selection. Finally, in the bar plot on the right, we examine the contribution of ensembling multiple captions. Using $\upsilon=3$ captions is approximately optimal. Note that the 7 individual captions exhibit high variance in zero-shot accuracy (under the ICE framework); this variance in performance is greatly reduced by ensembling a small number of captions.

\section{Limitations}
While our results show consistent improvements when evaluating on several different baseline methods across a diverse collection of datasets, we note several limitations of our method. 
First, as elaborated in Section \ref{subsec::methodology::caption_properties}, ICE heavily relies on the assumption that the captions can provide useful information about the image that is not fully encoded by the image embeddings.
As seen in Section \ref{subsec::experiments::analysis_on_ice}, when the captions provide unhelpful or adversarial information and there lacks a good selection of caption scores weight $\lambda$, ICE could decrease the base classification performance.
Second, determining a good choice for $\lambda$ is non-trivial, as it requires the selector to have an understanding for when to trust the caption or image scores more on a per-datapoint basis. 
This task is comparable with relevant literature on failure mode prediction using confidence estimation \citep{tsiligkaridis2020dirichlet, hendrycks2016misclassified, zhu2023rethinking}, which is known to be challenging. 
Finally, generating captions can be expensive, since each additionally generated caption requires an additional forward pass on the base model.
For image-text foundation models like CoCa \citep{yu2022coca}, which typically contain hundreds of millions of parameters, this strategy can quickly become a time bottleneck for ICE.
Thus, it is important to find a balance between the robustness benefits reaped from the number of captions used, and the linearly-increasing time costs for each additional caption generated.

\section{Conclusion}
We explored the use of captioners which is understudied in the vision-language literature for classification.
We proposed a novel method for improved zero-shot classification performance based on combining image and caption embeddings. 
We showed that our method can be easily paired with existing SOTA methods, and provide improvements of $0.5\%$ on average and up to $3\%$ across a diverse array of datasets in cross-dataset generalization and domain generalization.
Several ablation studies are presented to study sensitivity of parameters on performance.
We performed an in-depth analysis on why our method can help reclassify previously misclassified points, and also cover cases where it might fail.
Future work includes extending this work to better balance the weights between image and caption scores, and considering ways to generate more informative captions for improved downstream classification.

\section*{Acknowledgments}
DISTRIBUTION STATEMENT A. Approved for public release. Distribution is unlimited.

This material is based upon work supported by the Under Secretary of Defense for Research and Engineering under Air Force Contract No. FA8702-15-D-0001. Any opinions, findings, conclusions or recommendations expressed in this material are those of the author(s) and do not necessarily reflect the views of the Under Secretary of Defense for Research and Engineering.

\section*{Impact Statement}
This paper presents work whose goal is to advance the field of Machine Learning. There are many potential societal consequences of our work, none which we feel must be specifically highlighted here.

\bibliography{main}
\bibliographystyle{icml2024}

\newpage
\appendix
\onecolumn

\section{Experiment Implementation Details} \label{sec::exp_imp_details}
\paragraph{ICE.} 
For each method ICE is paired with, we simply take the image probability distribution output by our baseline method, and apply ICE with the computed caption scores. The hyperparameters we use to implement ICE in all experiments are $K=5$, $\xi = 0.08$, $\epsilon = 1 \times 10^{-12}$, $\upsilon = 3$, and $P = \{\text{``a''}, \text{``a photo of''}, \text{``a photo containing''}\}$. We find the best empirical performance when using the centroid of the embeddings of $3$ differently prompted captions and dynamically computing the caption scores weight $\lambda$ using Equation \ref{eq::lambda}.

\paragraph{Baselines.}
We make a good-faith attempt to tune the hyperparameters of each few-shot baseline. We use batch size 64 and SGD with momentum. Training data is sampled in a round robin fashion to maximize class diversity within each mini-batch. CLIPood trains the vision encoder for 750 iterations at learning rate $1\times 10^{-5}$  with adaptive margin value of 0.1. CoOp trains 3 prompt tokens initialized with ``a photo of'' for 1250 iterations at learning rate $2\times 10^{-4}$ with cross-entropy loss. MaPLe trains 3 prompt tokens prepended to each of the first 3 layers on both encoders for 750 iterations at learning rate $1$ with cross-entropy loss. 


\setlength\tabcolsep{5 pt}
\begin{table}[t!]
\scriptsize
\centering
\begin{tabular}{lccccccccccccccccc}
\toprule
& \textbf{Source} & \multicolumn{11}{c}{\textbf{Cross-dataset Evaluation Targets}} & \multicolumn{5}{c}{\textbf{Domain Generalization Targets}} \\
\cmidrule(lr){2-2} \cmidrule(lr){3-13} \cmidrule(lr){14-18}

     & \rotatebox{90}{ INet } & \rotatebox{90}{ Caltech } & \rotatebox{90}{ Pets } & \rotatebox{90}{ Cars } & \rotatebox{90}{ Flowers } & \rotatebox{90}{ Food } & \rotatebox{90}{ Aircraft } & \rotatebox{90}{ SUN } & \rotatebox{90}{ DTD } & \rotatebox{90}{ EuroSAT } & \rotatebox{90}{ UCF } & \rotatebox{90}{ Average } & \rotatebox{90}{ INet-V2 } & \rotatebox{90}{ INet-Sketch } & \rotatebox{90}{ INet-A } & \rotatebox{90}{ INet-R } & \rotatebox{90}{ Average }\\
    \midrule
    CLIPood  & 76.6 & \textbf{97.2} & \textbf{94.3} & 92.7 & 77.5 & 87.3 & 37.4 & \textbf{74.3} & 60.3 & 59.5 & 75.2 & 75.6 & \textbf{69.5} & \textbf{64.9} & 56.6 & 88.7 & \textbf{69.9} \\
    \rowcolor{lightgray}
      + ICE  & \textbf{76.7} & \textbf{97.2} & 93.9 & \textbf{93.1} & \textbf{77.8} & \textbf{87.4} & \textbf{39.6} & \textbf{74.3} & \textbf{60.8} & \textbf{61.8} & \textbf{75.7} & \textbf{76.2} & 69.2 & \textbf{64.9} & \textbf{56.7} & \textbf{88.8} & \textbf{69.9} \\
    \midrule
    CoOp  & 76.4 & \textbf{97.4} & \textbf{93.9} & 93.2 & 77.0 & \textbf{87.6} & 39.0 & 73.4 & 59.2 & 61.2 & 74.7 & 75.7 & \textbf{69.0} & 63.3 & 55.2 & 87.7 & 68.8 \\
    \rowcolor{lightgray}
      + ICE  & \textbf{76.7} & 97.2 & 93.8 & \textbf{93.5} & \textbf{77.7} & \textbf{87.6} & \textbf{40.7} & \textbf{73.6} & \textbf{60.6} & \textbf{62.0} & \textbf{75.1} & \textbf{76.2} & \textbf{69.0} & \textbf{63.4} & \textbf{55.9} & \textbf{88.1} & \textbf{69.1} \\
    \midrule
    MaPLe  & 77.3 & \textbf{96.7} & \textbf{94.2} & 92.8 & 76.9 & \textbf{87.2} & 39.5 & 73.9 & 61.1 & 58.7 & 76.0 & 75.7 & \textbf{70.2} & \textbf{64.7} & 54.8 & 88.2 & 69.5 \\
    \rowcolor{lightgray}
      + ICE  & \textbf{77.5} & 96.6 & 94.1 & \textbf{93.0} & \textbf{77.1} & \textbf{87.2} & \textbf{41.4} & \textbf{74.3} & \textbf{61.3} & \textbf{59.6} & \textbf{76.8} & \textbf{76.1} & 70.1 & 64.6 & \textbf{55.1} & \textbf{88.6} & \textbf{69.6} \\
    \bottomrule
  \end{tabular}
  \caption{Comparison with \emph{few-shot baselines} in the \emph{cross-dataset evaluation} setting and the \emph{domain generalization} setting. The model is fine-tuned on ImageNet with three different methods and tested on 15 total target datasets. The average accuracies are calculated separately for the two settings following prior work. In all cases, we observe that evaluating with our ICE method provides consistent improvements. All methods use CoCa ViT-L/14.}
  \vspace{-1.5em}
  \label{tab:ft}
\end{table}
\setlength\tabcolsep{6 pt}




\setlength\tabcolsep{5 pt}
\begin{table*}[t!]
\scriptsize
\centering
\begin{tabular}{lccccccccccccccccc}
\toprule
&  & \multicolumn{11}{c}{\textbf{Cross-dataset Evaluation Targets}} & \multicolumn{5}{c}{\textbf{Domain Generalization Targets}} \\
 \cmidrule(lr){3-13} \cmidrule(lr){14-18}

     & \rotatebox{90}{ INet } & \rotatebox{90}{ Caltech } & \rotatebox{90}{ Pets } & \rotatebox{90}{ Cars } & \rotatebox{90}{ Flowers } & \rotatebox{90}{ Food } & \rotatebox{90}{ Aircraft } & \rotatebox{90}{ SUN } & \rotatebox{90}{ DTD } & \rotatebox{90}{ EuroSAT } & \rotatebox{90}{ UCF } & \rotatebox{90}{ Average } & \rotatebox{90}{ INet-V2 } & \rotatebox{90}{ INet-Sketch } & \rotatebox{90}{ INet-A } & \rotatebox{90}{ INet-R } & \rotatebox{90}{ Average }\\
    \midrule

    CoCa Zero-shot (Image)  & 75.1 & \textbf{97.6} & \textbf{93.8} & 92.7 & 77.3 & 87.5 & 36.8 & 73.6 & 57.2 & 58.5 & 73.4 & 74.8 & 67.5 & 63.5 & 53.8 & 87.0 & 68.0 \\
    CoCa Zero-shot (Caption) & 58.8 & 85.6 & 76.3 & 83.1 & 63.6 & 72.7 & \textbf{40.7} & 54.8 & 44.0 & 34.9 & 60.3 & 61.6 & 50.2 & 50.7 & 38.7 & 73.8 & 53.3 \\
    \rowcolor{lightgray}
      + ICE  & \textbf{75.6} & 97.1 & \textbf{93.8} & \textbf{93.0} & \textbf{78.0} & \textbf{87.7} & {38.3} & \textbf{74.0} & \textbf{59.3} & \textbf{61.3} & \textbf{74.3} & \textbf{75.7} & \textbf{67.8} & \textbf{64.0} & \textbf{54.4} & \textbf{87.5} & \textbf{68.4} \\
    \midrule

    BLIP-2 Zero-shot (Image)  & 73.8 & 94.6 & \textbf{93.6} & 76.9 & \textbf{79.4} & \textbf{90.9} & \textbf{32.8} & 68.0 & 52.7 & 56.2 & 74.7 & 72.0 & 68.0 & 57.9 & 68.3 & 85.5 & 69.9 \\
    BLIP-2 Zero-shot (Caption) & 44.8 & 87.3 & 29.8 & 51.3 & 49.9 & 56.2 & 7.5 & 50.0 & 46.3 & 39.4 & 61.9 & 48.0 & 42.2 & 47.0 & 42.5 & 75.6 & 51.8 \\
    \rowcolor{lightgray}
      + ICE & \textbf{74.8} & \textbf{95.5} & 93.1 & \textbf{78.6} & \textbf{79.4} & 90.4 & 32.6 & \textbf{70.3} & \textbf{54.7} & \textbf{56.3} & \textbf{75.4} & \textbf{72.6} & \textbf{68.8} & \textbf{60.0} & \textbf{69.6} & \textbf{88.0} & \textbf{71.6} \\

    \midrule
    
    LLaVA Zero-shot (Image)  & 73.8 & 94.6 & \textbf{93.6} & \textbf{76.9} & \textbf{79.4} & \textbf{90.9} & \textbf{32.8} & 68.0 & 52.7 & 56.2 & \textbf{74.7} & \textbf{72.0} & 68.0 & 57.9 & 68.3 & 85.5 & 69.9 \\
    LLaVA Zero-shot (Caption) & 42.4 & 85.4 & 19.8 & 17.8 & 16.8 & 43.8 & 4.9 & 49.0 & 41.5 & 52.7 & 56.8 & 38.8 & 37.5 & 38.2 & 38.7 & 66.8 & 45.3 \\
    \rowcolor{lightgray}
      + ICE & \textbf{74.1} & \textbf{95.1} & 93.0 & 76.0 & 78.2 & 90.6 & 32.3 & \textbf{69.0} & \textbf{53.1} & \textbf{57.4} & 74.6 & 71.9 & \textbf{68.3} & \textbf{58.4} & \textbf{69.0} & \textbf{86.5} & \textbf{70.5} \\

    \bottomrule
   
  \end{tabular}
  \caption{Experiments using a variety of multimodal captioners, showing that using ICE to combine the image and caption embeddings improves the overall zero-shot accuracy. CoCa ViT-L/14 uses its own pretrained image encoder, while BLIP-2 and LLaVA use the frozen CLIP ViT-L/14 model from \cite{radford2021learning}. }
  \vspace{-1.5em}
  \label{tab:zs-appendix}
\end{table*}
\setlength\tabcolsep{6 pt}

\paragraph{Caption prompts.} For all ICE results, we use a set of 3 prompt templates to generate 3 diverse captions per image. These captions are stored in our GitHub repository under the ``captions'' folder for easy reference. Caption prompts are model specific due to the difference in pretraining.

CoCa caption prompts are dataset-agnostic: $\{\text{``a''}, \text{``a photo of''}, \text{``a photo containing''}\}$. BLIP-2 and LLaVA are optimized for VQA, so we found that the best classification results are achieved by prompting with a question and answer format, using dataset-specific questions. These prompts are listed in Tables \ref{tab:blip-prompts} and \ref{tab:llava-prompts}

\begin{table}[t!]
\centering
\begin{tabular}{l l }
\toprule
Dataset & LLaVA Caption prompt \\
\midrule
ImageNet & ``What is in this photo?'' \\
Caltech & ``What is in this photo?'' \\
Pets    & ``What type of pet is in this photo?'' \\
Cars  & ``What type of car is in this photo?'' \\
Flowers & ``What type of flower is in this photo?'' \\
Food & ``What type of food is in this photo'' \\
Aircraft & ``What type of aircraft is in this photo?'' \\
SUN & ``What is in this photo?'' \\
DTD & ``Describe the texture in this photo.'' \\
EuroSAT & ``What type of land use is in this satellite photo?'' \\
UCF & ``What type of action is in this photo?'' \\
ImageNet  & ``Question: What is in this photo? Answer: A photo of '' \\
ImageNet-V2 & ``What is in this photo?'' \\
ImageNet-Sketch & ``What is in this photo?'' \\
ImageNet-A & ``What is in this photo?'' \\
ImageNet-R & ``What is in this photo?'' \\
\bottomrule
  \end{tabular}
  \caption{Prompts used to generate the first LLaVA caption. The other two captions are generated by using the phrases ``be specific'' or ``be concise'' in the prompt to elicit a diverse set of responses from the VL model. For example, the two other ImageNet prompts are: ``What is in this photo? Be specific.'' and ``What is in this photo? Be concise.''}
  \label{tab:blip-prompts}
\end{table}

\begin{table}[t!]
\centering
\begin{tabular}{l l }
\toprule
Dataset & BLIP-2 Caption prompt \\
\midrule
ImageNet & ``Question: What is in this photo? Answer: A photo of '' \\
Caltech & ``Question: What is in this photo? Answer: A photo of '' \\
Pets    & ``Question: What type of pet is in this photo? Answer: A photo of '' \\
Cars  & ``Question: What type of car is in this photo? Answer: A photo of '' \\
Flowers & ``Question: What type of flower is in this photo? Answer: A photo of '' \\
Food & ``Question: What type of food is in this photo? Answer: A photo of '' \\
Aircraft & ``Question: What type of aircraft is in this photo? Answer: A photo of '' \\
SUN & ``Question: What is in this photo? Answer: A photo of '' \\
DTD & ``Question: Describe the texture in this photo. Answer: A photo of '' \\
EuroSAT & ``Question: What type of land use is in this satellite photo? Answer: A photo of '' \\
UCF & ``Question: What type of action is in this photo? Answer: A photo of '' \\
ImageNet  & ``Question: What is in this photo? Answer: A photo of '' \\
ImageNet-V2 & ``Question: What is in this photo? Answer: A photo of '' \\
ImageNet-Sketch & ``Question: What is in this photo? Answer: A photo of '' \\
ImageNet-A & ``Question: What is in this photo? Answer: A photo of '' \\
ImageNet-R & ``Question: What is in this photo? Answer: A photo of '' \\
\bottomrule
  \end{tabular}
  \caption{Prompts used to generate the first BLIP-2 caption. The other two captions are generated by using the phrases ``be specific'' or ``be concise'' in the prompt to elicit a diverse set of responses from the VL model. For example, the two other ImageNet prompts are: ``Question: What is in this photo? Be specific. Answer: A photo of '' and ``Question: What is in this photo? Be concise. Answer: A photo of ''}
  \label{tab:llava-prompts}
\end{table}

\end{document}